\documentclass{article}
\usepackage{amssymb}

\usepackage[final]{corl_2025} 

\usepackage{graphicx}
\usepackage{multirow}
\usepackage{array}
\usepackage{amssymb}
\usepackage{bbding}
\usepackage{booktabs}   
\usepackage{diagbox}
\usepackage{tabularx} 
\usepackage{ragged2e} 
\usepackage{enumitem}

\title{AgentWorld: An Interactive Simulation Platform for Scene Construction and Mobile Robotic Manipulation}

\author{
  Yizheng Zhang\textsuperscript{1*}, Zhenjun Yu\textsuperscript{2*}, Jiaxin Lai\textsuperscript{1}, Cewu Lu\textsuperscript{2}, Lei Han\textsuperscript{1 \S}\\
  \textsuperscript{1}Tencent Robotics X, \textsuperscript{2}Shanghai Jiao Tong University \\
  \textsuperscript{1}\tt\small zyz.robotics@gmail.com, jesselai@tencent.com, leihan.cs@gmail.com \\
  \textsuperscript{2}\tt\small \{jeffson-yu, lucewu\}@sjtu.edu.cn \\
}

\begin{document}
\maketitle

\vspace{-0.5cm}
\begin{figure*}[!ht]
    \centering
    \includegraphics[width=1.0\linewidth]{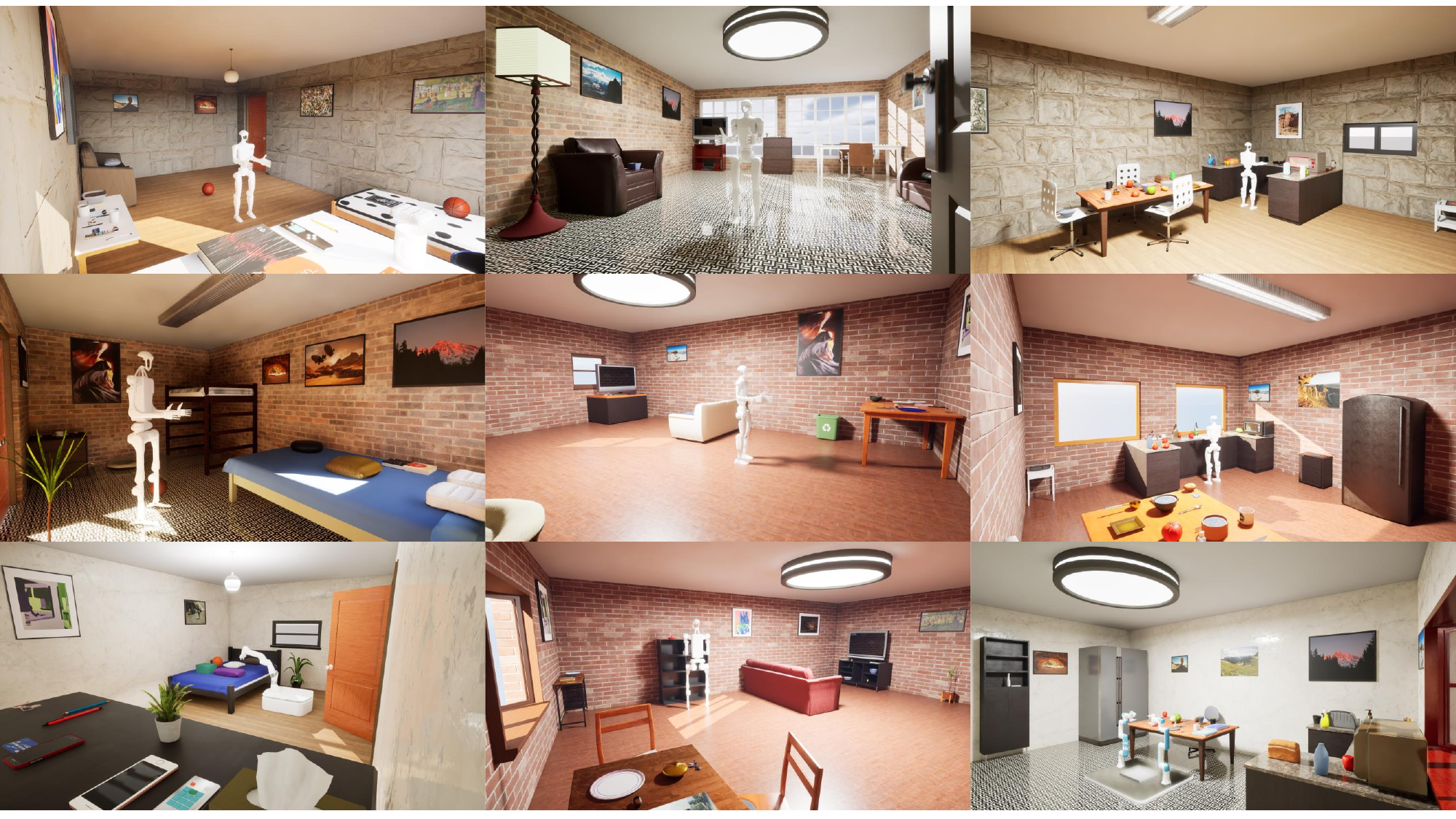}
    \caption{\textbf{Overview of AgentWorld.} AgentWorld simulation platform features several core abilities for embodied AI: (1) Procedural scene construction supporting various layout generation. (2) Abundant semantic 3D assets repository with realistic visual material and physical properties. (3) Mobile-based Teleoperation system for robotic manipulation.}
    \label{fig:teaser}
\end{figure*}

\begin{abstract}
We introduce \textit{AgentWorld}, an interactive simulation platform for developing household mobile manipulation capabilities. Our platform combines automated scene construction that encompasses layout generation, semantic asset placement, visual material configuration, and physics simulation, with a dual-mode teleoperation system supporting both wheeled bases and humanoid locomotion policies for data collection. The resulting \textit{AgentWorld Dataset} captures diverse tasks ranging from primitive actions (pick-and-place, push-pull, etc.) to multistage activities (serve drinks, heat up food, etc.) across living rooms, bedrooms, and kitchens. Through extensive benchmarking of imitation learning methods including behavior cloning, action chunking transformers, diffusion policies, and vision-language-action models, we demonstrate the dataset's effectiveness for sim-to-real transfer. The integrated system provides a comprehensive solution for scalable robotic skill acquisition in complex home environments, bridging the gap between simulation-based training and real-world deployment. The code, datasets will be available at \url{https://yizhengzhang1.github.io/agent_world/}
\end{abstract}

\renewcommand{\thefootnote}{}
\footnotetext{\noindent * indicates equal contributions. \S \ Lei Han is the corresponding author.}



\section{Introduction}\label{sec:intro}



Recent advancements in embodied AI and robotic manipulation have highlighted the need for scalable, interactive simulation environments that support both scene construction and data collection for training autonomous agents. While existing platforms \citep{ai2thor, procthor, ssg, rcaregen} offer partial solutions, such as scene generation\citep{infiniteworld, ssg, rcaregen} or task-specific manipulation datasets \citep{mimicgen, bigym, humanoidbench}, few provide a unified framework that integrates high-fidelity scene construction with flexible mobile robotic data collection system. To bridge this gap, we present \textbf{AgentWorld}, an interactive simulation platform designed for \textbf{procedural scene construction} and \textbf{mobile-based teleoperation}, enabling efficient data collection for imitation learning in complex household environments.

AgentWorld addresses two critical challenges in embodied AI research: (1) \textbf{stable and diverse scene generation}, ensuring that the simulated environments are visually realistic and physically plausible, and (2) \textbf{a comprehensive data collection system} in simulation, which allows seamless control of mobile bases and robotic arms for data collection. AgentWorld is built upon NVIDIA's Omniverse Isaac Sim \citep{isaacsim} and Unreal Engine \citep{UE}, allowing it to inherit both strengths including the physics engine for robot parallel training and realistic rendering effects.

For procedural construction of diverse household scenes, our platform supports four key components: (i) \textbf{Layout Generation}, which procedurally generates different room layouts with walls, stairs and floors; (ii) \textbf{Semantic Asset Selection and Placement}, leveraging a rich 3D asset repository and arranging objects in semantically meaningful configurations; (iii) \textbf{Visual Material Configuration}, enabling fine-grained semantic object material appearances; and (iv) \textbf{Interactive Physics Simulation}, ensuring dynamic interactions of robots and assets remain stable and accurate.

Beyond scene construction, AgentWorld introduces a \textbf{Mobile-based Data Collection system} that combines keyboard-controlled mobile base navigation with arm and hand manipulation. For facilitating the collection of diverse robotic interaction data for both wheel-based and legged robots, we implement a reinforcement learning method for lower-limb control of humanoid robots. A VR-based teleoperation system is employed for arm control and hand keypoint detection, and we use a retargeting approach for operating both grippers and dexterous hands.

We validate our platform by constructing the AgentWorld Dataset, in which we generate 150 household scenes with more than 9000 assets and 4 different embodiments with either gripper or dexterous hand as end effectors. Utilizing our data collection system, we collect more than 1000 robot manipulation trajectories with tasks from simple object interactions to long-horizon, multistage challenges. Using imitation learning algorithms (BC, ACT\citep{act}, Diffusion Policy\citep{dp}, and $\pi_0$ \citep{pi0}), we demonstrate successful policy training and sim-to-real transfer with few-shot real data finetuning, highlighting the utility of our dataset for real-world robotic applications.

Our primary contributions include:

(1) A procedural scene construction framework supporting diverse layout generation and semantic object placement, features by an extensive 3D asset library with realistic visual and physical properties, enhancing simulation fidelity.

(2) A mobile teleoperation system for data collection, enabling precise control of both mobile bases and manipulators with grippers and dexterous hands.

(3) The AgentWorld Dataset, a large-scale benchmark for robotic manipulation, was validated through imitation learning and sim-to-real experiments.


\section{Related Work}\label{sec:related_work}
\paragraph{Simulation Platform for Embodied AI}
Recent progress in embodied AI has been driven by simulation platforms that balance realism, scalability, and task diversity. \citep{RFUniverse, rcareworld} establish robotic simulation environments with Unity and extend the Python-Unity communication interface, while still lacking programmatic scene construction and accurate physics simulation. \citep{procthor} employs procedural generation to create large-scale interactive environments, while \citep{maniskill, maniskill2} focus on manipulation tasks with physically realistic interactions. Similarly, \citep{rlbench} provides a benchmark for robotic learning with a suite of predefined tasks, and \citep{behavior1k} extends scene complexity by simulating human-like activities. \citep{grutopia} introduces a social interaction layer, enabling NPC-driven task allocation, and \citep{robocasa} emphasizes household robotics with large-scale object interactions. While these platforms excel in specific domains, AgentWorld distinguishes itself by integrating procedural scene construction with mobile teleoperation, offering a more holistic approach to embodied AI research.

\paragraph{Scene Construction}
Effective scene construction requires both geometric diversity and physical realism. Prior works \citep{infiniteworld, ssg, ai2thor, rcaregen} have explored procedural generation with primitive methods or large language models. However, many existing systems rely on static assets with limited material customization. SceneCAD \citep{scenecad} introduces a data-driven approach for indoor scene synthesis, while 3DTopia-XL \citep{3dtopiaxl} leverages generative models for diverse 3D asset creation. In contrast, AgentWorld introduces a dynamic asset repository with PBR materials, ensuring realistic visual and physical behaviors. Our platform also supports automated layout generation, allowing users to define semantic rules for object arrangement.

\paragraph{Data Collection for robotic manipulation}
Data collection methodologies in robotics have evolved from scripted demonstrations \citep{dexcap} to human teleoperation  \citep{mobilealoha}. While \citep{act} synthesizes manipulation data via motion retargeting, it lacks mobile base integration. AgentWorld advances this domain by combining keyboard-controlled mobility with arm and hand teleoperation, enabling the collection of long-horizon, multistage tasks. Our approach aligns with recent trends in imitation learning \citep{dp, dp3} while addressing the sim-to-real gap through physics-aware simulation.


\section{AgentWorld Simulation Platform}\label{sec:method}

\begin{table}[h]
\centering
\resizebox{14cm}{!}{

\begin{tabular}{c|ccc|ccc|ccc}
\toprule
\multirow{3}{*}{Name} & \multicolumn{3}{c}{\textbf{Asset}} & \multicolumn{3}{c}{\textbf{Robotic Platforms}} & \multicolumn{3}{c}{\textbf{Data Collection}} \\

\multirow{3}{*}{} & Num of    & Material  & Physics & \multirow{2}{*}{Fixed-B} & \multirow{2}{*}{Mobile-B} & \multirow{2}{*}{Legged} & Tele-     & Dexterous & Num of \\
\multirow{3}{*}{} & Assets & Selection & Config  & \multirow{2}{*}{       } & \multirow{2}{*}{        } & \multirow{2}{*}{      } & Operation & Hand      & Trajectories \\
\midrule

Maniskill2\citep{maniskill2}            & 2144     & \Checkmark        & \Checkmark       & \Checkmark   & \Checkmark   & \XSolidBrush & \XSolidBrush & \XSolidBrush & 30k  \\
ProcTHOR\citep{procthor}                & 3578     & \Checkmark        & \XSolidBrush     & \XSolidBrush & \Checkmark   & \XSolidBrush & \XSolidBrush & \XSolidBrush & --  \\
RLBench\citep{rlbench}                  & 28       & \XSolidBrush      & \Checkmark       & \Checkmark   & \XSolidBrush & \XSolidBrush & \XSolidBrush & \XSolidBrush & --    \\
BiGym\citep{bigym}                      & $<$200   & \XSolidBrush      & \Checkmark       & \Checkmark   & \Checkmark   & \Checkmark    & J + FB       & \XSolidBrush & $>$2000    \\
Behavior-1K\citep{behavior1k}           & 5215     & \Checkmark        & \Checkmark       & \XSolidBrush & \Checkmark   & \XSolidBrush & \XSolidBrush & \XSolidBrush & --  \\
MimicGen\citep{mimicgen}                & 40       & \XSolidBrush      & \Checkmark       & \Checkmark   & \Checkmark   & \Checkmark   & \XSolidBrush & \XSolidBrush & 50k    \\
RoboCasa\citep{robocasa}                & 2509     & \XSolidBrush      & \Checkmark       & \Checkmark   & \Checkmark   & \Checkmark   & J + FB       & \XSolidBrush & 100k    \\
InfiniteWorld\citep{infiniteworld}      & $>$10000 & \XSolidBrush      & \Checkmark       & \Checkmark   & \Checkmark   & \Checkmark   & \XSolidBrush & \XSolidBrush & --    \\
GRUtopia\citep{grutopia}                & $\approx$25000 & \XSolidBrush      & \Checkmark       & \Checkmark   & \Checkmark   & \Checkmark   & FB           & \XSolidBrush & --    \\
\midrule
AgentWorld                              & $>$9000  & \Checkmark        & \Checkmark       & \Checkmark    & \Checkmark   & \Checkmark   & J + FB + L   & \Checkmark   & $>$1000  \\
\bottomrule
\end{tabular}
}
\vspace{0.1cm}
\caption{Comparison of robotic simulation platforms in terms of asset properties, robotic platform support, and data collection capabilities. 
Fixed-B and Mobile-B stands for fixed and mobile base robots. The teleoperation colomn demonstrates support for joint action control (J), floating base control (FB), and locomotion control (L) for humanoid robots.
\textbf{AgentWorld} represents our proposed platform integrating all key capabilities.}
\label{tab:sim_compare}
\end{table}

We introduce the AgentWorld simulation platform, which mainly consists of two parts: \textbf{Procedural Scene Generation}, and \textbf{Data Collection with Teleoperation}. Our simulation system can programmatically construct various types of household scene with abundant 3D assets, and provides a VR-based data collection system for recording human-demonstrate manipulation data. We compare the AgentWorld simulation platform with popular platforms in Tab. \ref{tab:sim_compare}.

\subsection{Procedural Scene Generation}

\begin{figure*}[!ht]
    \centering
    \includegraphics[width=1.0\linewidth]{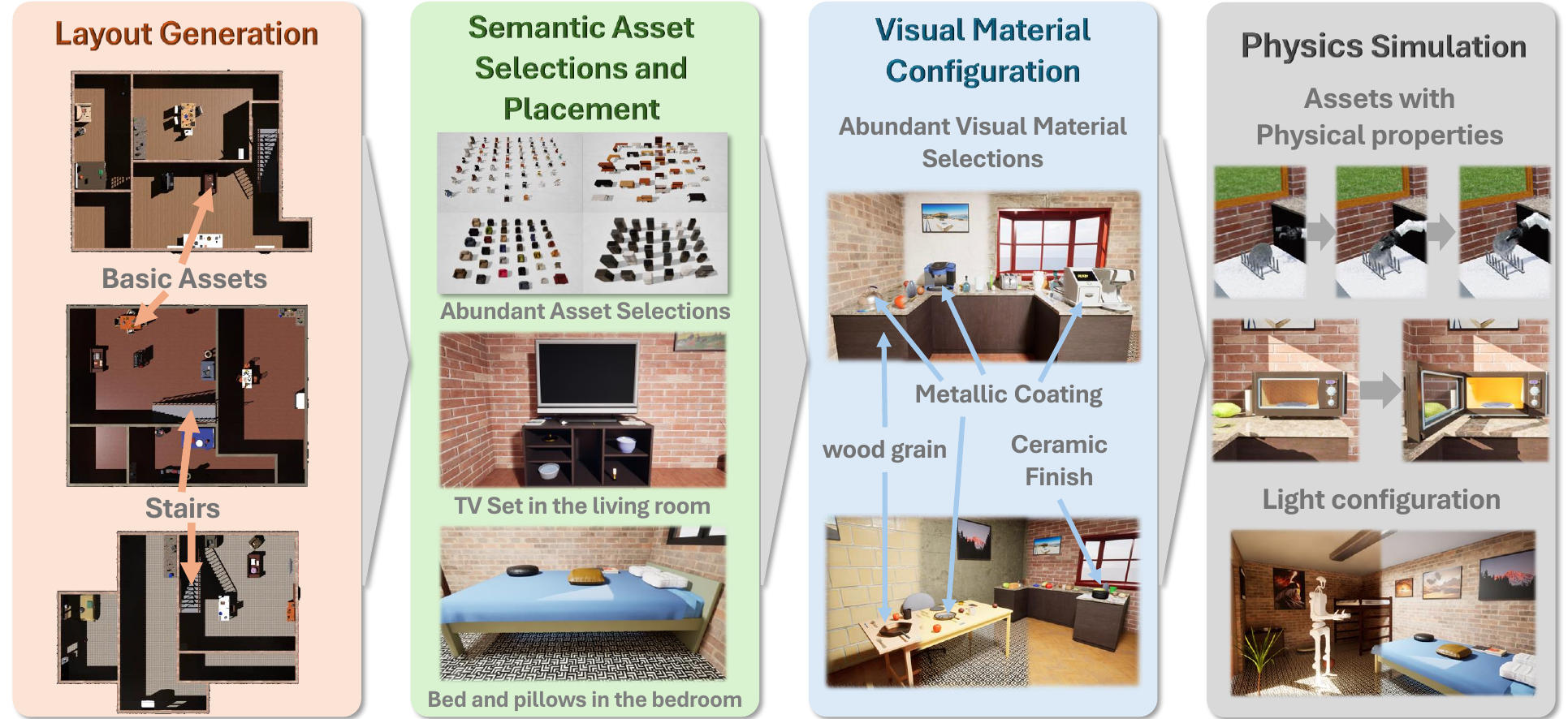}
    \caption{Pipeline of the scene construction module in AgentWorld.}
    \label{fig:scene_pipeline}
\end{figure*}

To construct large-scale, realistic simulation environments for household scenarios and tasks, we employ a programmatic approach with an interactive interface for scene generation. As illustrated in Fig. \ref{fig:scene_pipeline}, our pipeline mainly comprises four key stages: \textbf{Layout Generation}, \textbf{Semantic Asset Selections and Placement}, \textbf{Visual Material Configuration}, and \textbf{Interactive Physics Simulation}. The first three stages are implemented in Unreal Engine \citep{UE}, leveraging its superior rendering capabilities and intuitive blueprint system for efficient scene design. For the Interactive Physics Simulation stage, we utilize NVIDIA's Omniverse Isaac Sim \citep{isaacsim}, which provides enhanced physics engine performance crucial for sim-to-real transfer and enables efficient parallel training of robotic agents.

\subsubsection{Layout Generation}
The first stage of our pipeline generates complete room layouts based on user-specified room types. Our system currently supports three fundamental room categories: \textit{Living room}, \textit{Kitchen}, and \textit{Bedroom}. The procedural generation module automatically constructs architectural elements including walls, ceilings, and floors, with support for both single-room configurations and multi-room combinations. Additionally, as demonstrated in Fig. \ref{fig:scene_pipeline}, our framework can generate multi-floor environments interconnected by staircases.

Following the architectural layout, the system populates the scene with category-appropriate functional assets such as furniture and decorative elements. The asset selection mechanism and its underlying methodology are detailed in the following section.

\subsubsection{Semantic Asset Selections and Placement}
To enable diverse scene construction, our simulation platform integrates a comprehensive collection of open-source 3D assets sourced from repositories including ProcTHOR~\citep{procthor} and Behavior-1K~\citep{behavior1k}. Each asset has been manually annotated with semantic metadata describing its functional properties. We classify these assets into two primary categories: \textbf{Basic Assets} and \textbf{Interactable Assets}.

\textbf{Basic Assets} constitute the fundamental elements of room layouts and are automatically generated during the initial scene construction phase with contextually appropriate placements. This category encompasses room-specific furnishings, such as sofas and TV sets in the living room, beds and chairs in the bedroom, and tables in the kitchen, etc.

\textbf{Interactable Assets} represent objects that serve functional purposes and can be manipulated by robotic agents. Our scene generation module supports dynamic addition and adjustment of these assets during the second construction phase. We further distinguish several object types: articulated assets like microwaves, refrigerators and closets that have doors, and rigid objects such as household tools (knives, broomsticks, etc.) and common items like fruits, bottles and toys, etc. Leveraging this extensive interactable asset repository enables the simulation of numerous household manipulation tasks, which we detail in Section~\ref{sec:task}.

Our system incorporates intelligent placement algorithms that position assets according to their semantic functions (e.g., placing food items on dining surfaces, pillows on bedding). The interactive interface provides fine-grained control over asset positioning and orientation adjustments.

\subsubsection{Visual Material Configuration}
To better facilitate the sim-to-real paradigm, our platform incorporates a diverse selection of visual materials to enhance scene generalization through effective data augmentation. We implement an extensive library of high-fidelity Physically Based Rendering (PBR) materials covering various categories of common household surfaces. As shown in Fig.~\ref{fig:scene_pipeline}, our system supports material customization at multiple levels: for architectural components like walls and floors, we provide realistic material options such as marble and brick textures; for 3D assets, we enable property-appropriate material adjustments including wood grains, ceramic finishes, fabric textures, and metallic coatings. This comprehensive material system allows for precise visual customization while maintaining physical accuracy in rendering, significantly expanding the variety of achievable visual appearances for robust robotic training.

\subsubsection{Interactive Physics Simulation}
The Interactive Physics Simulation module implements AgentWorld's robotic interaction layer using NVIDIA Isaac Sim with its GPU-accelerated PhysX 5.0 engine. Our system automatically configures physical properties for all 3D assets based on semantic annotations, generating optimized collision primitives (convex hulls, convex decomposition or SDF meshes, etc.) and assigning material properties including friction coefficients, restitution values, and mass distributions. 

For articulated mechanisms such as doors and drawers, the system precisely configure joint parameters including: joint types (revolute for cabinets/doors, prismatic for drawers), movement ranges, material-specific friction coefficients (wood: $0.4\pm 0.1$, metal: $0.2\pm 0.05$), and actuation models. The simulation environment incorporates Isaac Sim's advanced lighting system, providing photometrically accurate illumination ($50$ - $20,000$ lux) with programmable color temperature ($2700$K - $6500$K) and light exposure ($-5.0$ - $5.0$) through various light sources.

\subsection{Mobile-based Teleoperation for data collection}
\begin{figure*}[!ht]
    \centering
    \includegraphics[width=1.0\linewidth]{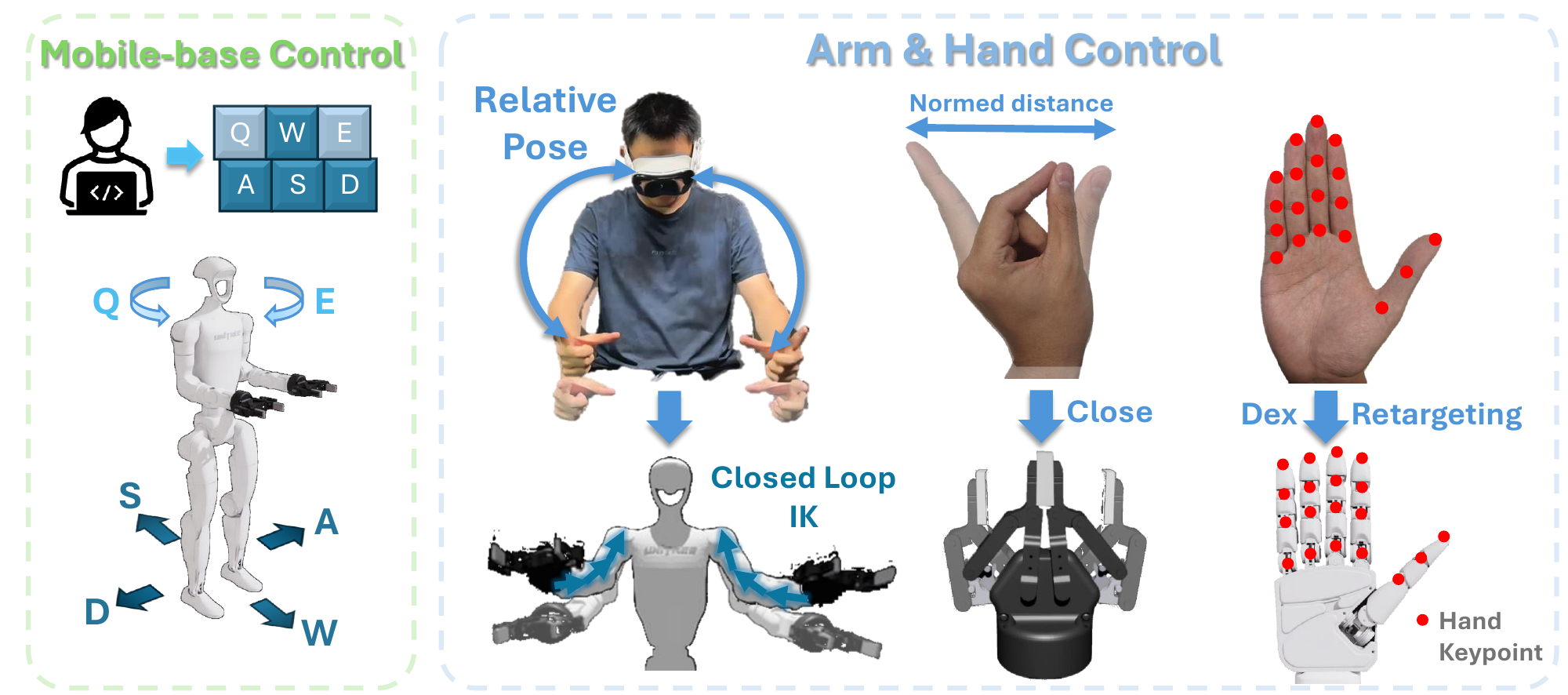}
    \caption{Data collection system of AgentWorld. For the mobile-base control, we allow the users to use the keyboard to to control robots, both wheel-based and legged. For arm \& hand control, we use the VR head set to get the hand pose and compute IK for obtaining the arm action, and utilize re-targeting methods to drive robotic hands.}
    \label{fig:teleop_pipeline}
    \vspace{-0.5cm}
\end{figure*}

Teleoperation serves as an effective method for collecting robotic manipulation data, though it presents particular challenges for long-horizon tasks in expansive household environments. These challenges primarily stem from the need for coordinated mobile navigation between multiple locations. Our teleoperation system addresses these difficulties through the integrated pipeline shown in Fig.~\ref{fig:teleop_pipeline}, which comprises two key components: \textbf{Mobile-base Control} and \textbf{Arm \& Hand Control}.

\paragraph{Mobile-base Control.} In order to realize mobile manipulation tasks data collection, we aim to make the mobile-based robot move around in the room. Basically, for every time step $t$, we tend to control the robot with an action $a_t = ( v_x, v_y, v_\theta ) \in \mathcal{A}$ with $\mathcal{A} = \mathbb{R}^3$ a 3 DOFs action space , where $v_x$ and $v_y$ is the robot velocity of the movement on the plane, and $v_\theta$ is the velocity of the yaw angle of the robot floating base.

For wheeled or direct controllable floating base robots, movement control is straightforward. For legged humanoids, we employ IsaacLab's reward-driven locomotion policy~\citep{mittal2023orbit}, adapted for dual-arm stability. The model enables robot movement control via keyboard as demonstrated in Fig. \ref{fig:teleop_pipeline}. We can switch the robot between locomotion and manipulation by fixing or freeing its base link. The training details will be discussed in the supplementary material.

\paragraph{Arm \& Hand Control.} We implement the OpenTeleVision system \citep{opentelevision} in IsaacLab to control robotic arms with grippers or dexterous hands. The VR headset captures hand keypoints in its coordinate frame. With the headset camera fixed on the head of a humanoid robot (or a certain reasonable location for other embodiments), we calibrate the relative pose between hand wrists and headset in real world to match the camera-end effector relationship in simulation. Arm joint poses are computed using a Closed-loop Inverse Kinematics (CLIK) algorithm based on Pinocchio \citep{pinocchio}.

For different types of end effectors, as shown in Fig. \ref{fig:teleop_pipeline}, the gripper action is controlled by normalizing the thumb-to-index fingertip distance, and we employ dex-retargeting \citep{anyteleop} to convert hand keypoints to joint actions of the dexterous hands.


\section{AgentWorld Dataset}\label{sec:dataset}
Based on our AgentWorld simulation platform, we construct a comprehensive dataset containing diverse household activities performed by different embodiments. This section details our task settings across different scene layouts and the collected dataset specifications.

\subsection{Scene \& Task Description}\label{sec:task}
We focus on daily household activities involving both bimanual and single-arm manipulation tasks. Our environment features three primary room layouts: \textbf{Living room}, \textbf{Bedroom}, and \textbf{Kitchen}. For each room type, we generate 10 distinct base layouts using our scene generation module. After establishing the basic structure, we randomly populate each room with inter-actable 3D assets with reasonable placements, ultimately creating 150 unique room configurations. To cover most of the household activities, we separate our task settings in two kinds: \textbf{Basic Tasks} and \textbf{Multistage Tasks}.

\textbf{Basic Tasks.} We collect sequences of data on basic manipulation tasks in order to enable robots to learn some basic human operations, including but not limited to:
\begin{itemize}[leftmargin=0.7cm]
    \item \textbf{Pick \& Place}: Simple pick \& place tasks are basic ability for robots to master. In our dataset, we control the gripper jaw or dexterous hand of the robot to pick cups, fruits, plates and toys, etc, and place the objects in boxes, sinks, trash bins or racks.
    \item \textbf{Open \& Close}: We teleoperate the robot to conduct articulated manipulation tasks, including opening doors of refrigerators, cabinets, and the ovens, followed by closing actions to bring articulation knowledge to manipulation models.
    \item \textbf{Push \& Pull}: The dataset includes manipulation tasks where the robot interacts with drawers (pulling them open or pushing them shut), slides objects (such as boxes or books) across tables, and operates buttons or switches, to teach precise position controls to the robot.
\end{itemize}

\textbf{Multistage Tasks.}
After introducing basic tasks of the robots, we believe it is crucial for the robots to learn how to perform household activities that are more meaningful. Therefore, we design and record multistage tasks in three typical home environments:
\begin{itemize}[leftmargin=0.7cm]
    \item \textbf{Living Room}: Tasks include \textit{Organizing books} (put the books on shelves), \textit{serving drinks} (pick up the pitcher and the cup and pour the drink), and \textit{cleaning tables} (put trash in trash cans).
    \item \textbf{Bedroom}: We simulate activities like \textit{making a bed} (Put the objects on the bed away, placing pillows), \textit{organizing a wardrobe} (open/close closets, hang the clothes rack), and \textit{setting an alarm clock} (pressing buttons or sliding switches).
    \item \textbf{Kitchen}: Complex interactions such as \textit{Store the food} (pick up food on the table and store it into the fridge), \textit{heat up the food} (pick up food and put it into the microwave, and then press the button), and \textit{Clean the dishes} (use a sponge to clean the dishes and place them into a rack).
\end{itemize}
These tasks require sequential decision-making and combine primitive skills (pick \& place, Open \& Close and push \& pull) to achieve higher-level goals, mimicking real-world household demands.

\vspace{-0.2cm}
\subsection{Dataset Details}\label{sec:data_detail}
We employ four different embodiments:  Unitree G1 \citep{G1}, H1 \citep{H1}, Franka Emika panda with a wheel base, and a DOBOT X-Trainer \citep{xtrainer} with a fixed platform. The end effectors of the two humanoid robots are set to be either the 2F-85 Gripper \citep{2f85} or the TRX-Hand5 \citep{trxhand}.
For the basic tasks, we have provided 10 different assets of every task category, and recorded 10 sequences of each robot manipulating a certain asset with different poses, with the length of the sequences from 80 to 150 steps. As for the multistage tasks, we recorded 30 to 50 sequences for every task, with the length from 300 to 800 steps. The control frequency is fixed to 30 fps. In our settings, a \textbf{Wrist Camera} is attached on each end effector on the robot arms, and a \textbf{Head Camera} is set immobile on the humanoid robots. For the X-Trainer, two \textbf{Wrist Cameras} and a fixed \textbf{Third-person Camera} are set to align with the real-world setting. For each frame, we record the current robot joint poses, the action to take, and RGB-D output of the cameras. More dataset details will be discussed in the supplementary material.  

\vspace{-0.2cm}
\section{Experiments}

\subsection{Imitation Learning for AgentWorld Dataset}

To comprehensively evaluate our AgentWorld dataset, we benchmark multiple imitation learning paradigms: (1) conventional \textit{Behavior Cloning (BC)} as our baseline; (2) \textit{Action Chunking Transformers (ACT)} \citep{act} for sequence-aware action prediction; (3) \textit{Diffusion Policies (DP)} \citep{dp} for stochastic action modeling; and (4) $\pi_0$ \citep{pi0} that integrates language instructions with visual inputs. All models process synchronized multi-view RGB images, with $\pi_0$ additionally using language task descriptions. The unified output space comprises: (1) dual-arm joint and end effector actions, (2) 3-DOF velocity commands for mobile base navigation, and (3) a binary locomotion-manipulation mode selector that determines whether to prioritize mobile navigation or stationary manipulation. The velocity commands will be forwarded to the mobile-base controller in our data collection system, either a wheel-base driver or a pretrained legged locomotion policy.

\begin{figure*}[!ht]
    \centering
    \includegraphics[width=1\linewidth]{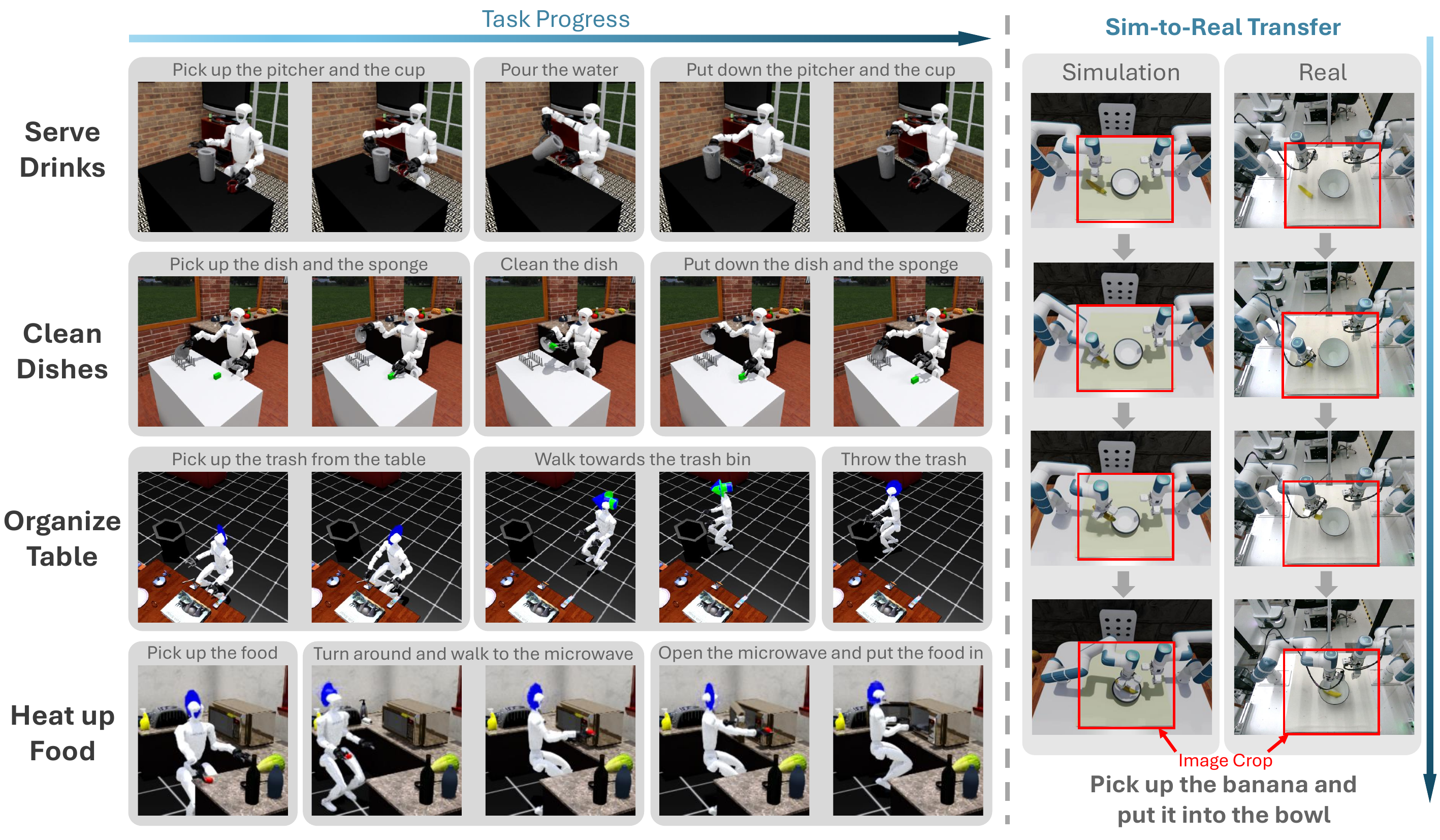}
    \caption{Qualitative results for different imitation learning algorithms in AgentWorld Dataset, and a Sim-to-real transfer example to validate the availability and generalizability of our data.}
    \label{fig:qual}
    \vspace{-0.4cm}
\end{figure*}

For Basic Tasks, we train per-category models and evaluate with unseen assets/surroundings. For Multistage Tasks, we train on single tasks and evaluate with varied asset poses and materials. We evaluate 50 episodes of each task with every model, and use success rate to assess the models' performances. 
The average success rates of every task category are demonstrated in Tab.~\ref{tab:quan}. For basic manipulation tasks, ACT \cite{act} achieves the most consistent performance (62-76\% success rate) due to its action chunking mechanism effectively handling short-horizon action sequences. In multistage tasks, $\pi_0$ \cite{pi0} shows superior results (20-30\% v.s. others' 4-16\%), benefiting from its pretrained representation that captures long-horizon task structures. However, all methods struggle with complex mobile manipulation, where both precise control and long-term planning are simultaneously required.
Several demonstrations for the inference results of imitation learning algorithms are shown in Fig. \ref{fig:qual}. The training details will be elaborated in our supplementary material.

\begin{table*}[t!]
    \center
    \small
    \begin{tabular}{c|c|c|c|c|c|c}
        \toprule
        \multirow{2}{*}{\diagbox{Algo.}{Task}} & \multicolumn{3}{c|}{Basic Tasks} & \multicolumn{3}{c}{Multistage Tasks} \\ 
        \multirow{2}{*}{} & Pick \& Place & Open \& Close & Push \& Pull & Living Room & Kitchen & Bedroom \\
        \midrule

        BC                 & 52\%          & 62\%           & 58\%          & 10\%          &  4\%          &  6\%          \\
        ACT\citep{act}     & \textbf{66\%} & \textbf{84\%}  & 72\%          & 16\%          & 12\%          & 14\%          \\
        DP\citep{dp}       & 64\%          & 78\%           & \textbf{76\%} & 28\%          & \textbf{20\%} & \textbf{24\%} \\
        $\pi_0$\citep{pi0} & 64\%          & 82\%           & 70\%          & \textbf{30\%} & \textbf{20\%} & 18\%          \\

        \bottomrule

    \end{tabular}
    \caption{Quantitative results for different imitation learning algorithms in AgentWorld Dataset.}
    \label{tab:quan}
    \vspace{-0.4cm}
\end{table*}

\subsection{Real-world Experiment}\label{sec:real}
To validate the sim-to-real transfer capability of our platform, we conducted real-world basic task experiments involving placing various objects into a bowl. We trained the $\pi_0$ policy using a sim-real-combined paradigm: initial training on simulation data (10 assets with 10 trajectories each) followed by fine-tuning with limited real-world demonstrations (3 objects - banana, apple, mug - with 3 trajectories each with varied object positions). This combined approach achieved a \textbf{29.3\%} success rate, and as illustrated in Fig.~\ref{fig:qual}, the policy successfully handles basic manipulation tasks despite the reality gap. These results suggest that AgentWorld's realistic material rendering and physical simulation properties provide an effective foundation for sim-to-real transfer, while highlighting the importance of targeted real-world fine-tuning for practical deployment.

\section{Conclusion}\label{sec:conclusion}
We present AgentWorld, a versatile simulation platform that bridges the gap between procedural scene construction and mobile robotic data collection for embodied AI research, which addresses key challenges in robotic learning, demonstrated through successful imitation learning experiments and sim-to-real validation. Our experiments with the AgentWorld Dataset show that the platform's flexible scene configuration and comprehensive teleoperation system enable effective training of manipulation policies. Future work will expand the platform's capabilities to include more complex physical interactions and multi-agent scenarios, further enhancing its utility for robotics research.

\section{Limitations}\label{sec:limit}
Despite the comprehensive capabilities of AgentWorld, we identify two key technical limitations in the current implementation:

\begin{itemize}[leftmargin=0.7cm]
    \item \textbf{Limited Deformable Object Support}: Due to constraints in the underlying Isaac Sim physics engine, our platform currently has limited support for simulating deformable objects (e.g., cloth, ropes, or soft bodies). This restricts the range of manipulation tasks that can be accurately modeled and transferred to real-world scenarios involving such objects.
    \item \textbf{Pure Sim-to-Real Realization}: The significant domain gap between simulation and reality requires extensive manual tuning of rendering parameters to achieve visual alignment. While we have demonstrated successful sim-to-real transfer for simple tasks through simulation pretraining followed by few-shot real-world fine-tuning, it's still difficult for real-world deployment with policies trained on pure simulation data. Previous work\cite{maddukuri2025simandreal} has shown an improvement in real-world performance when combined synthetic data with real data, and Isaac Sim’s validated physics and rendering assure the effectiveness of our platform. 
 Besides, multistage manipulation tasks still demand substantially more real-world demonstrations to bridge the reality gap effectively. Future works include verifying the effectiveness of the introduction of the simulation data, realizing sim-to-real transfer for long-horizon tasks based on pure synthetic data, etc.
\end{itemize}

These limitations point to important future research directions, including integration of more advanced physics engines and development of robust domain adaptation techniques for challenging manipulation scenarios.

\clearpage

\bibliography{example}  

\clearpage

\appendix

\section{Supplementary Video}
The supplementary video presented in our website provides dynamic visualizations of AgentWorld's core capabilities, organized as follows:

\begin{itemize}[leftmargin=0.7cm]
    \item \textbf{Scene Construction Demonstration (0:00-0:45)}:
    
    \begin{itemize}
        \item \textbf{0:00-0:13}: Layout generation showcasing different room organization with stairs.
        \item \textbf{0:13-0:20}: Semantic asset selection and placement demonstration.
        \item \textbf{0:20-0:29}: Visual material configuration with walls, floors and assets with various materials.
        \item \textbf{0:29-0:45}: Interactive physics simulation demonstrating inter-actable assets with physics properties like collisions and joints, etc.
    \end{itemize}
    
    \item \textbf{Teleoperation System (0:45-2:04)}:
    \begin{itemize}
        \item \textbf{0:45-1:05}: An example of VR-based arm and gripper teleoperation.
        \item \textbf{1:05-2:04}: An example of dexterous hand control with manipulation of objects and articulated assets.
    \end{itemize}
    
    \item \textbf{Learning Results (2:04-2:56)}:
    \begin{itemize}
        \item \textbf{2:04-2:42}: Imitation learning performance in simulation for multistage tasks.
        \item \textbf{2:42-2:56}: A Sim-to-real transfer demonstration of simple pick \& place task.
    \end{itemize}
\end{itemize}

The video complements our paper by providing real-time demonstrations of AgentWorld's interactive features and experimental outcomes. Each segment highlights key functionalities of our simulation platform.

\section{Details on Procedural Scene Generation}

Our procedural scene generation framework is \textbf{fully automatic} for scenes constructed from existing assets, requiring no manual intervention for object placement or physics configuration. For new assets, only \textbf{minimal one-time human input} is needed during initialization: assigning categorical labels, room types, and optional co-occurrence constraints (following ProcTHOR\citep{procthor} design principles). Subsequent bounding box calculations and physics property assignments (e.g., mass, friction) are handled automatically. With this approach, unlimited scenes can be generated without additional human effort once assets are annotated.

The system extends ProcTHOR's rule-based algorithms by integrating Unreal Engine's \textbf{Procedural Content Generation (PCG)} framework, enabling real-time scene generation (2--3 seconds per scene) with visual feedback. Key technical improvements include:

\begin{itemize}[leftmargin=*,noitemsep]
    \item \textbf{Geometric realism}: Wall thickness simulation and optimized room layouts using metrics like aspect ratio, floor-to-bounding-box area ratio, and L-shape expansion to produce natural, rectangular spaces.
    \item \textbf{Layout pragmatism}: Enhanced inter-room connectivity checks to avoid impractical designs (e.g., isolated rooms) and support for multi-floor structures.
    \item \textbf{Physics automation}: Rule-based material properties (e.g., ``wood'' $\rightarrow$ predefined mass/friction) triggered by semantic asset tags.
\end{itemize}

While scene generation and physics are fully automatic after asset setup, the initial \textbf{semantic tagging of new assets} (e.g., labeling a chair as ``furniture'') remains the sole manual step. These enhancements collectively advance environmental realism beyond ProcTHOR's original implementation while maintaining computational efficiency.

\section{Training Details}

\subsection{Locomotion Policy for Humanoid Robots}
As shown in Table \ref{tab:rewards}, we define a series of rewards to encourage humanoid robots to achieve walking with different terrain adaptation such as flat planes and stairs of varying steepness, etc.
Meanwhile, we define relevant termination conditions to end a certain current episode, as shown in Table \ref{tab:termination}.
In addition, we use the foot trajectory generator\citep{lee2020learning} to generate a reference trajectory for robot feet. The output of the locomotion policy network will be added as an increment to this reference trajectory. In this way, we can narrow the exploration range of the action and accelerate the training process.
During the training process, we randomly set different joystick control commands within a certain period, enabling the robot to encounter as many different control commands as possible. The goal of the training is to make the robot follow these commands to train the robot's locomotion policy.

\begin{table}[htbp]
    \centering
    
    \label{tab:rewards} 
    \begin{tabularx}{\textwidth}{lX}
        \toprule
        Reward & \RaggedRight Description \\
        \midrule
        track\_angle\_vel\_z\_exp & Encourages the robot to follow the rotation commands in the yaw direction. \\
        track\_linear\_vel\_xy\_exp & Encourages the robot to follow the translation commands in the x and y directions. \\
        linear\_vel\_z\_l2 & Penalizes the robot for fluctuations in the z - direction. \\
        feet\_air\_time & Encourages the robot to lift its feet rather than drag them on the ground. \\
        feet\_slide & Penalizes the robot for foot sliding. \\
        dof\_acc\_l2 & Penalizes the robot's joint accelerations to ensure smooth joint movements. \\
        dof\_torques\_l2 & Penalizes the robot's joint torques to keep the joint torques as stable as possible. \\
        dof\_pos\_limits & Penalizes the robot when its joints exceed the defined limits. \\
        joint\_deviation\_arms & Penalizes the deviation of the robot's arm joints from their default positions, guiding the policy to explore around the default positions. \\
        joint\_deviation\_hip & Penalizes the deviation of the robot's hip joints from their default positions, guiding the policy to explore around the default positions. \\
        joint\_deviation\_torso & Penalizes the deviation of the robot's torso joints from their default positions, guiding the policy to explore around the default positions. \\
        \bottomrule
    \end{tabularx}
    \vspace{0.1cm}
    \caption{Rewards}
\end{table}

\begin{table}[tbp]
    \centering
    
    \label{tab:termination}
    \begin{tabularx}{\textwidth}{lX}
        \toprule
        Condition & \RaggedRight Description \\
        \midrule
        base\_contact & If the robot's body collides with the ground, it indicates that the robot has fallen. \\
        time\_out & Limits the maximum time for each training episode. \\
        \bottomrule
    \end{tabularx}
    \vspace{0.1cm}
    \caption{Termination Conditions}
    \vspace{-0.5cm}
\end{table}

In practical applications, the robot's upper limbs will grasp different objects, which means that the positions of the upper limbs can vary widely. Therefore, to ensure that our locomotion policy is robust enough, we have incorporated randomization of the arm positions during the training process. Based on the given default arm position, for every 150 frames, we superimpose the result of a uniform sampling as the arm target joint position. This approach enables the training to cover a wide range of different positions of the upper limbs.

\subsection{Implementation of Imitation Learning Algorithms}

\paragraph{Observation \& Action Space.} The input RGB observations maintain a consistent resolution of $480 \times 640$. The robot's proprioceptive state is defined as $s = [s_{qpos}, s_e, s_f]$, where $s_{qpos}$ represents joint positions of the arms, $s_e$ denotes end effector poses, and $s_f = (x, y, \theta)$ is the floating base's plane location and yaw angle. As previously described, the algorithm outputs the action $a = [a_{qpos}, a_e, a_f, 0/1]$, where $a_{qpos}$ and $a_e$ specify arm joint and end effector actions, $a_f = (v_x, v_y, v_\theta)$ represents mobile base velocity commands, and $0/1$ the binary value selects between locomotion and manipulation modes. The Unitree G1, H1, and Franka Emika Panda arm each possess $7$ degrees of freedom (DOFs), while the dual-arm X-Trainer has $6$ DOFs per arm. For end effectors, grippers are controlled via a single normalized value for opening distance, and the TRX-Hand5 features $18$ actuated DOFs.
\paragraph{Algorithm Architecture.} Our behavior cloning algorithm utilizes a ResNet-18 \citep{resnet} image encoder for visual feature extraction. The flattened image features are concatenated with proprioception states and processed through a 5-layer MLP for action prediction. Layer normalization and tanh activation are applied to the output. For ACT, Diffusion Policy, and $\pi_0$, we adopt the LeRobot \citep{lerobot} implementation with action chunk sizes of $20$ for basic tasks and $40$ for multistage tasks. All approaches incorporate \textit{temporal ensemble} to enhance action smoothness.

\paragraph{Implementation Details.} All algorithms are trained with batch size $16$, except $\pi_0$ which uses $8$. Training spans 50K steps for basic tasks and 150K for multistage tasks. We observe convergence across all methods with 80K-100K steps, with no subsequent performance increases. On NVIDIA L20 GPU, training durations are 6 hours for behavior cloning, 12 hours for ACT and Diffusion Policy, and 18 hours for $\pi_0$.

\section{Dataset Details}
Table. \ref{tab:dataset} demonstrates detailed information of our datasets including concrete descriptions and number of trajectories of each task in every category. Our dataset currently contains 1,150 trajectories ("$>1000$" indicates active expansion). We're actively expanding the collection to improve generalization and sim-to-real transfer, and our platform supports community contributions to scale diversity.

\begin{table}[htbp]
    \centering
    \renewcommand\arraystretch{1.5}
    \label{tab:dataset} 
    \begin{tabularx}{\textwidth}{lXl}
        \toprule
        Category & Task Description & Num of Trajectories \\
        \midrule
        \multirow{4}{*}{Pick \& Place} & Pick some objects and place them in the bowl. & $10\times10$ \\
        \multirow{4}{*}{             } & Pick some objects and place them into the opened microwave / fridge / stove. & $30\times10$ \\
        \multirow{4}{*}{             } & Pick some objects out from the bowl.  & $10\times10$ \\
        \multirow{4}{*}{             } & Pick some objects out from the opened microwave / fridge / stove. & $30\times10$ \\
        \hline
        \multirow{2}{*}{Open \& Close} & Open \& Close doors of rooms. & $10\times20$ \\
        \multirow{2}{*}{             } & Open \& Close doors of the furniture: fridges, microwaves, stoves. & $30\times20$ \\
        \hline
        \multirow{3}{*}{Push \& Pull}  & Pull out \& Push in the drawers. & $10\times20$ \\
        \multirow{3}{*}{            }  & Push the buttons on microwaves / stoves. & $10\times20$ \\ 
        \multirow{3}{*}{            }  & Push some objects on the tables away from the robot.& $10\times10$ \\ 
        
        \midrule
        \multirow{3}{*}{Living Room} & Pick up the books on the table and walk to the shelf and organize them. & $10\times50$ \\ 
        \multirow{3}{*}{           } & Pick up the pitcher and the cup and pour the drink. & $5\times30$ \\ 
        \multirow{3}{*}{           } & Pick up the trash on the table and walk to the trash bin to throw them away & $10\times50$ \\ 
        \hline
        
        \multirow{3}{*}{Bedroom} & Slide the objects on the bed away and put the pillow on the top of the bed & $10\times50$ \\ 
        \multirow{3}{*}{       } & Pick up the clothes rack and open the closets and hang them in. & $10\times50$ \\ 
        \multirow{3}{*}{       } & Pick up the alarm clock and push the button. & $5\times30$ \\ 
        \hline
        
        \multirow{3}{*}{Kitchen}  & Pick up dishes on the table and store it into the fridge. & $5\times30$ \\ 
        \multirow{3}{*}{       }  & Pick up the food and turn around to open the microwave and put the food in. & $5\times50$ \\ 
        \multirow{3}{*}{       }  & Pick up the dishes on the rack and the sponge and clean the dishes. & $5\times50$ \\ 
        
        \bottomrule
    \end{tabularx}
    \vspace{0.1cm}
    \caption{Dataset details for task descriptions and number of trajectories. The number of trajectories is demonstrated as number of assets $\times$ number of recorded sequences.}
    
\end{table}

\clearpage



\end{document}